\documentclass[a4paper,12pt]{article}

\usepackage{fancyvrb}
\usepackage{graphicx}
\usepackage{hyperref}
\usepackage{amssymb}

\begin{document}

\title{Information compression, intelligence, computing, and mathematics}

\author{J Gerard Wolff\footnote{Dr Gerry Wolff, BA (Cantab), PhD (Wales), CEng, MBCS (CITP); CognitionResearch.org, Menai Bridge, UK; \href{mailto:jgw@cognitionresearch.org}{jgw@cognitionresearch.org}; +44 (0) 1248 712962; +44 (0) 7746 290775; {\em Skype}: gerry.wolff; {\em Web}: \href{http://www.cognitionresearch.org}{www.cognitionresearch.org}.}}

\maketitle

\begin{abstract}

This paper presents evidence for the idea that much of artificial intelligence, human perception and cognition, mainstream computing, and mathematics, may be understood as compression of information via the matching and unification of patterns. This is the basis for the {\em SP theory of intelligence}, outlined in the paper and fully described elsewhere. Relevant evidence may be seen: in empirical support for the SP theory; in some advantages of information compression (IC) in terms of biology and engineering; in our use of shorthands and ordinary words in language; in how we merge successive views of any one thing; in visual recognition; in binocular vision; in visual adaptation; in how we learn lexical and grammatical structures in language; and in perceptual constancies. IC via the matching and unification of patterns may be seen in both computing and mathematics: in IC via equations; in the matching and unification of names; in the reduction or removal of redundancy from unary numbers; in the workings of Post's Canonical System and the transition function in the Universal Turing Machine; in the way computers retrieve information from memory; in systems like Prolog; and in the query-by-example technique for information retrieval. The chunking-with-codes technique for IC may be seen in the use of named functions to avoid repetition of computer code. The schema-plus-correction technique may be seen in functions with parameters and in the use of classes in object-oriented programming. And the run-length coding technique may be seen in multiplication, in division, and in several other devices in mathematics and computing. The SP theory resolves the apparent paradox of ``decompression by compression''. And computing and cognition as IC is compatible with the uses of redundancy in such things as backup copies to safeguard data and understanding speech in a noisy environment.

\end{abstract}


\section{Introduction}


\begin{quote}

``Fascinating idea! All that mental work I've done over the years, and what have I got to show for it? A goddamned zipfile! Well, why not, after all?'' (John Winston Bush, 1996).

\end{quote}

This paper describes a range of observations and arguments in support of the idea that much of artificial intelligence, human perception and cognition, mainstream computing, and mathematics, may be understood as compression of information via the matching and unification of patterns. These observations and arguments provide the foundation for the {\em SP theory of intelligence} and its realisation in the {\em SP computer model}---outlined below and described more fully elsewhere---in which information compression is centre stage. The aim here is to review, update, and extend the discussion in \cite{wolff_1993}, itself the basis for \cite[Chapter 2]{wolff_2006}.

\subsection{Related ideas}

Related ideas have been around from at least as far back as the 14th century when William of Ockham suggested that if something can be explained by two or more rival theories, we should choose the simplest. Later, Isaac Newton wrote that ``Nature is pleased with simplicity'' \cite[p.~320]{newton_2014}; Ernst Mach \cite{banks_2004} and Karl Pearson \cite{pearson_1892} suggested independently that scientific laws promote ``economy of thought''; Albert Einstein wrote that ``A theory is more impressive the greater the simplicity of its premises, the more different things it relates, and the more expanded its area of application.''\footnote{Quoted in \cite[p.~512]{isaacson_2007}}; cosmologist John Barrow has written that ``Science is, at root, just the search for compression in the world'' \cite[p.~247]{barrow_1992}; and George Kingsley Zipf \cite{zipf_1949} developed the idea that human behaviour is governed by a ``principle of least effort''.

Partly inspired by the publication of Claude Shannon's ``theory of communication'' \cite{shannon_1948} (now called ``information theory''), Fred Attneave \cite{attneave_1954}, Horace Barlow \cite{barlow_1959,barlow_1969} and others examined the role of information compression (IC) in the workings of brains and nervous systems.

The close connection between information compression and several other inter-related topics has been demonstrated by several researchers including Ray Solomonoff (inductive inference \cite{solomonoff_1964} and `algorithmic probability theory' \cite{solomonoff_1997}), Chris Wallace (classification \cite{wallace_boulton_1968} and inference \cite{wallace_freeman_1987}), Jorma Rissanen (modelling by shortest description \cite{rissanen_1978} and `stochastic complexity' \cite{rissanen_1987}), Andrey Kolmogorov and Gregory Chaitin (`algorithmic information theory' (see, for example, \cite{li_vitanyi_2009}), and Satosi Watanabe (pattern recognition \cite{watanabe_1972}). And Ray Solomonoff \cite{solomonoff_1986} has argued that the great majority of problems in science and mathematics may be seen as either `machine inversion' problems or `time limited optimization' problems, and that both kinds of problem can be solved by inductive inference using the principle of `minimum length encoding'.

In later research, information compression has featured as a guiding principle for artificial neural networks (see, for example, \cite{schmidhuber_2015}, Section 4.4)  and in research on grammatical inference (see, for example, \cite{sakamoto_2014}).

\subsection{Novelty and contribution}

The ideas described in this paper provide a perspective on artificial intelligence, human perception and cognition, mainstream computing, and mathematics, which is not widely recognised. The main features distinguishing it from other research are:

\begin{itemize}

\item The scope is very much broader than it is, for example, in the previously-mentioned research on artificial neural networks or grammatical inference. The thrust of the paper is evidence pointing to information compression via the matching and unification of patterns as an organising principle across diverse aspects of artificial intelligence, human perception and cognition, mainstream computing, and mathematics.

\item Most research relating to information compression and its applications makes extensive use of mathematics. By contrast, information compression in this paper and in the SP theory focusses on the simple primitive idea, dubbed ``ICMUP'' and described in Section \ref{preliminaries_section}, that redundancy in information may be reduced by finding patterns that match each other and merging or unifying patterns that are the same. Far from using mathematics as a basis for understanding information compression, the paper argues, in Section \ref{computing_mathematics_section}, that ICMUP may provide a basis for mathematics.

\item Although this is not the main focus of the paper, it is pertinent to mention that ICMUP provides the foundation for the distinctive and powerful concept of {\em multiple alignment}, a central part of the SP theory of intelligence and, on evidence to date, a key to versatility and adaptability in intelligent systems.

\end{itemize}

I believe this perspective is important for the field of artificial intelligence for three main reasons:

\begin{itemize}

\item It has things to say directly about the nature of perception, learning, and other aspects of artificial intelligence.

\item It provides a foundation for the SP theory of intelligence which, via the SP computer model, has demonstrable capabilities in several aspects of artificial intelligence, as outlined in Section \ref{outline_sp_theory_section}, and it has a range of potential benefits and applications ({\em ibid.}).

\item It suggests how artificial intelligence may be developed within an encompassing theoretical framework that includes human perception and cognition, mainstream computing, and mathematics.

\end{itemize}

\subsection{Apparent contradictions and their resolution}

Given that large amounts of information can be produced by people, by computers, and via mathematics, and given that `redundancy' or repetition in information is often useful in the storage and processing of information, it may seem perverse to suggest that IC is fundamental in our thinking, or in computing and mathematics. But for reasons outlined in Section \ref{resolving_apparent_contractions_section}, these apparent contradictions can be resolved.

\subsection{Presentation}

As an introduction to what follows, the next section describes some basic principles of IC. After that, the SP theory is described in outline, with pointers to further sources of information, and a summary of empirical support for the theory. This last is itself evidence for the importance of IC in computing and cognition. The sections that follow describe several other strands of evidence that point in the same direction.

\section{Preliminaries: information compression via the matching and unification of patterns}\label{preliminaries_section}

To cut through some of the complexities in this area, I have found it useful to focus on a rather simple idea: that we may identify repetition or `redundancy' in information by searching for patterns that match each other, and that we may reduce that redundancy and thus compress information by merging or `unifying' two or more copies to make one. For the sake of brevity, this idea may be shortened to ``information compression via the matching and unification of patterns'' or ``ICMUP''.

As just described, ICMUP loses information about the positions of all but one of the original patterns. But this can be remedied with any of the three variants of the idea, described below.

\subsection{Concepts of number}

ICMUP may seem too trivial to deserve comment. But because it is the foundation on which the rest of the SP system is built, there are implications that may seem strange and may at first sight look like major shortcomings in the theory:

\begin{itemize}

\item The first of these is that the SP system, in itself, has no concepts of number and has no procedures for processing numbers. Unlike an ordinary operating system or programming language, there is no provision for integers or reals and no functions such as addition, subtraction, square roots, or the like.\footnote{As we shall see in Section \ref{outline_sp_theory_section}, the SP system does use a concept of frequency and it does calculate probabilities. But these are part of the workings of the system and not available to users. In any case, they may be modelled via analogue signals, without using conventional concepts of number or arithmetic.}

\item Secondly, because the system has no concepts of number, it does not use any of the compression techniques that depend on numbers, such as arithmetic coding, wavelet compression, Huffman codes, or the like.

\end{itemize}

Although the core of the SP system lacks any concept of number, there is potential for the system to represent numbers and process them, provided that it is supplied with knowledge about Peano's axioms and related information about the structure and functioning of numbers, as outlined in \cite[Chapter 10]{wolff_2006}. The potential advantage of starting with a clean slate, focussing on the simple `primitive' concept of ICMUP, is that it can help us avoid old tramlines, and open doors to new ways of thinking.

\subsection{Variants of ICMUP}

With the first variant of ICMUP---a technique called {\em chunking-with-codes}---the unified pattern is given a relatively short name, identifier, or `code' which is used as a shorthand for the pattern or `chunk'. If, for example, the words ``Treaty on the Functioning of the European Union'' appear in several different places in a document, we may save space by writing the expression once, giving it a short name such as ``TFEU'', and then using that name as a code or shorthand for the expression wherever it occurs. Likewise for the abbreviations in this paper, ``IC'' and ``ICMUP''.

\sloppy Another variant, {\em schema-plus-correction}, is like chunking-with-codes but the unified chunk of information may have variations or `corrections' on different occasions. For example, a six-course menu in a restaurant may have the general form `\texttt{Menu1:~Appetiser (S) sorbet (M) (P) coffee-and-mints}', with choices at the points marked `\texttt{S}' (starter), `\texttt{M}' (main course), and `\texttt{P}' (pudding). Then a particular meal may be encoded economically as something like `\texttt{Menu1:(3)(5)(1)}', where the digits determine the choices of starter, main course, and pudding.

A third variant, {\em run-length coding}, may be used where there is a sequence two or more copies of a pattern, each one except the first following immediately after its predecessor. In this case, the multiple copies may be reduced to one, as before, with something to say how many copies there are, or when the sequence begins and ends, or, more vaguely, that the pattern is repeated. For example, a sports coach might specify exercises as something like ``touch toes ($\times 15$), push-ups ($\times 10$), skipping ($\times 30$), ...'' or ``Start running on the spot when I say `start' and keep going until I say `stop'\thinspace''.

\section{Outline of the SP theory of intelligence}\label{outline_sp_theory_section}

The {\em SP theory of intelligence}, described most fully in \cite{wolff_2006} and more briefly in \cite{sp_extended_overview}, aims to simplify and integrate observations and concepts across artificial intelligence, human perception and cognition, mainstream computing, and mathematics, with ICMUP as a unifying theme.

The theory, as it stands now is the product of an extended programme of development and testing via the SP computer model. It is envisaged that that model will be the basis for a high-parallel, open-source version of the {\em SP machine}, hosted on an existing high-performance computer, and accessible via the web. This will be a means for researchers everywhere to explore what can be done with the system and to create new versions of it \cite[Section 3.2]{sp_extended_overview}, \cite{sp_proposal}.

The SP theory, via the SP computer model, has demonstrable capabilities in areas that include the representation of diverse forms of knowledge (including class hierarchies, part-whole hierarchies, and their seamless integration), unsupervised learning, natural language processing, fuzzy pattern recognition and recognition at multiple levels of abstraction, best-match and semantic forms of information retrieval, several kinds of reasoning (one-step `deductive reasoning', abductive reasoning, probabilistic networks and trees, reasoning with `rules', nonmonotonic reasoning, explaining away, causal reasoning, and reasoning that is not supported by evidence), planning, problem solving, and information compression \cite{wolff_2006,sp_extended_overview}. It also has useful things to say about aspects of neuroscience and of human perception and cognition ({\em ibid.}).

Several potential benefits and applications of the SP theory are described in \cite{sp_benefits_apps}, with more detail in \cite{sp_vision} (understanding natural vision and the development of articial vision), \cite{sp_big_data} (how the SP theory may help to solve nine problems associated with big data), \cite{sp_autonomous_robots} (the development of computational and energy efficiency, of versatility, and of adaptability in autonomous robots), \cite{wolff_sp_intelligent_database} (the SP system as an intelligent database), and \cite{wolff_medical_diagnosis} (application of the SP system to medical diagnosis). An introduction to the theory may be seen in \cite{sp_in_brief}.

In broad terms, the SP theory has three main elements:

\begin{itemize}

\item All kinds of knowledge are represented with {\em patterns}: arrays of atomic symbols in one or two dimensions.

\item At the heart of the system is compression of information via the matching and unification (merging) of patterns, and the building of {\em multiple alignments} like the two shown in Figure \ref{fruit_flies_figure}.\footnote{The example sentence is the second part of {\em Time flies like an arrow. Fruit flies like a banana.}, attributed to Groucho Marks.} Here, the concept of multiple alignment has been borrowed and adapted from bioinformatics.

\item The system learns by compressing {\em New} patterns to create {\em Old} patterns like those shown in rows 1 to 8 in each of the two multiple alignments in the figure.

\end{itemize}

\begin{figure}[!htbp]
\fontsize{07.00pt}{08.40pt}
\centering
{\bf
\begin{BVerbatim}
0               fruit        flies            like              a        banana           0
                  |            |               |                |          |
1          A 12 fruit #A       |               |                |          |              1
           |          |        |               |                |          |
2     NP 2 A          #A N     |   #N #NP      |                |          |              2
      |                  |     |   |   |       |                |          |
3     |                  N 7 flies #N  |       |                |          |              3
      |                                |       |                |          |
4     |                                |       |                |    N 5 banana #N        4
      |                                |       |                |    |          |
5     |                                |       |      NP 3 D    | #D N          #N #NP    5
      |                                |       |      |    |    | |                 |
6     |                                |  V 9 like #V |    |    | |                 |     6
      |                                |  |        |  |    |    | |                 |
7 S 1 NP                              #NP V        #V NP   |    | |                #NP #S 7
                                                           |    | |
8                                                          D 11 a #D                      8

(a)

0         fruit        flies                 like                a        banana                0
            |            |                    |                  |          |
1           |            |                    |             D 11 a #D       |                   1
            |            |                    |             |      |        |
2           |            |                    |        NP 3 D      #D N     |    #N #NP         2
            |            |                    |        |              |     |    |   |
3           |            |                    |        |              N 5 banana #N  |          3
            |            |                    |        |                             |
4     N 6 fruit #N       |                    |        |                             |          4
      |         |        |                    |        |                             |
5 S 0 N         #N V     |   #V ADP           |        |                             |  #ADP #S 5
                   |     |   |   |            |        |                             |   |
6                  |     |   |   |    ADV 15 like #ADV |                             |   |      6
                   |     |   |   |     |           |   |                             |   |
7                  |     |   |  ADP 4 ADV         #ADV NP                           #NP #ADP    7
                   |     |   |
8                  V 8 flies #V                                                                 8

(b)
\end{BVerbatim}
}
\caption{Two multiple alignments created by the SP computer model showing two different parsings of the ambiguous sentence {\em Fruit flies like a banana}. Adapted from Figure 5.1 in \protect\cite{wolff_2006}, with permission.}
\label{fruit_flies_figure}
\end{figure}

Because information compression is intimately related to concepts of prediction and probability \cite{li_vitanyi_2009}, the SP system is fundamentally probabilistic. Each SP pattern has an associated frequency of occurrence, and probabilities may be calculated for multiple alignments and for inferences drawn from multiple alignments \cite[Section 4.4]{sp_extended_overview}, \cite[Section 3.7]{wolff_2006}. Although the system is fundamentally probabilistic, it may be constrained to deliver all-or-nothing results in the manner of conventional computing systems.

Since IC is central in the SP theory, the descriptive and explanatory range of the theory is itself evidence in support of the proposition that IC is a central principle in human perception and thinking, in computing and in mathematics.

\section{Biology and engineering}

This section and those that follow describe other evidence for the importance of IC in computing and cognition. First, let's take a bird's eye view of why IC might be important in people and other animals, and in computing.

In terms of biology:

\begin{itemize}

\item IC can confer a selective advantage to any creature by allowing it to store more information in a given storage space or use less space for a given amount of information, and by speeding up transmission of information along nerve fibres---thus speeding up reactions---or reducing the bandwidth needed for any given volume of information.

\item Perhaps more important than any of these things is the close connection, already mentioned, between IC and inductive inference. Compression of information provides a means of predicting the future from the past and estimating probabilities so that, for example, an animal may get to know where food may be found or where there may be dangers.

    Incidentally, the connection between IC and inductive prediction makes sense in terms of the matching and unification of patterns: any repeating pattern---such as the association between black clouds and rain---provides a basis for prediction---black clouds suggest that rain may be on the way---and probabilities may be derived from the number of repetitions.

\item Being able to make predictions and estimate probabilities can mean large savings in the use of energy with consequent benefits in terms of survival.

\end{itemize}

As with living things, IC can be beneficial in computing---in terms of the storage and transmission of information and what is arguably the fundamental purpose of computers: to make predictions. It may also have a considerable impact in increasing the energy efficiency of computers \cite[Section IX]{sp_big_data}, \cite[Section III]{sp_autonomous_robots}.

As we shall see, IC is more widespread in ordinary computers than may superficially appear. 

\section{Hiding in plain sight}

Compression of information is so much embedded in our thinking, and seems so natural and obvious, that it is easily overlooked. Here are some examples.

\subsection{Words as codes or shorthands}

In the same way that ``TFEU'' is a convenient code or shorthand for ``Treaty on the Functioning of the European Union'', a name like ``New York'' is a compact way of referring to the many things of which that renowned city is composed. Likewise for the many other names that we use: ``Nelson Mandela'', ``George Washington'', ``Mount Everest'', and so on.

More generally, most words in our everyday language stand for {\em classes} of things and, as such, are powerful aids to economical description. Imagine how cumbersome things would be if, on each occasion that we wanted to refer to a ``table'', we had to say something like ``A horizontal platform, often made of wood, used as a support for things like food, normally with four legs but sometimes three, ...'', like the slow language of the Ents in Tolkien's {\em The Lord of the Rings}. Likewise for verbs like ``speak'' or ``dance'', adjectives like ``artistic'' or ``exuberant'', and adverbs like ``quickly'' or ``carefully''.\footnote{Although natural language provides a very effective means of compressing information about the world, it is not free of redundancy. And that redundancy has a useful role to play in, for example, enabling us to understand speech in noisy conditions, and in learning the structure of language \cite[Section 5.2]{sp_extended_overview}.}

\subsection{Merging multiple views to make one}

\begin{figure}[!htbp]
\centering
\includegraphics[width=0.9\textwidth]{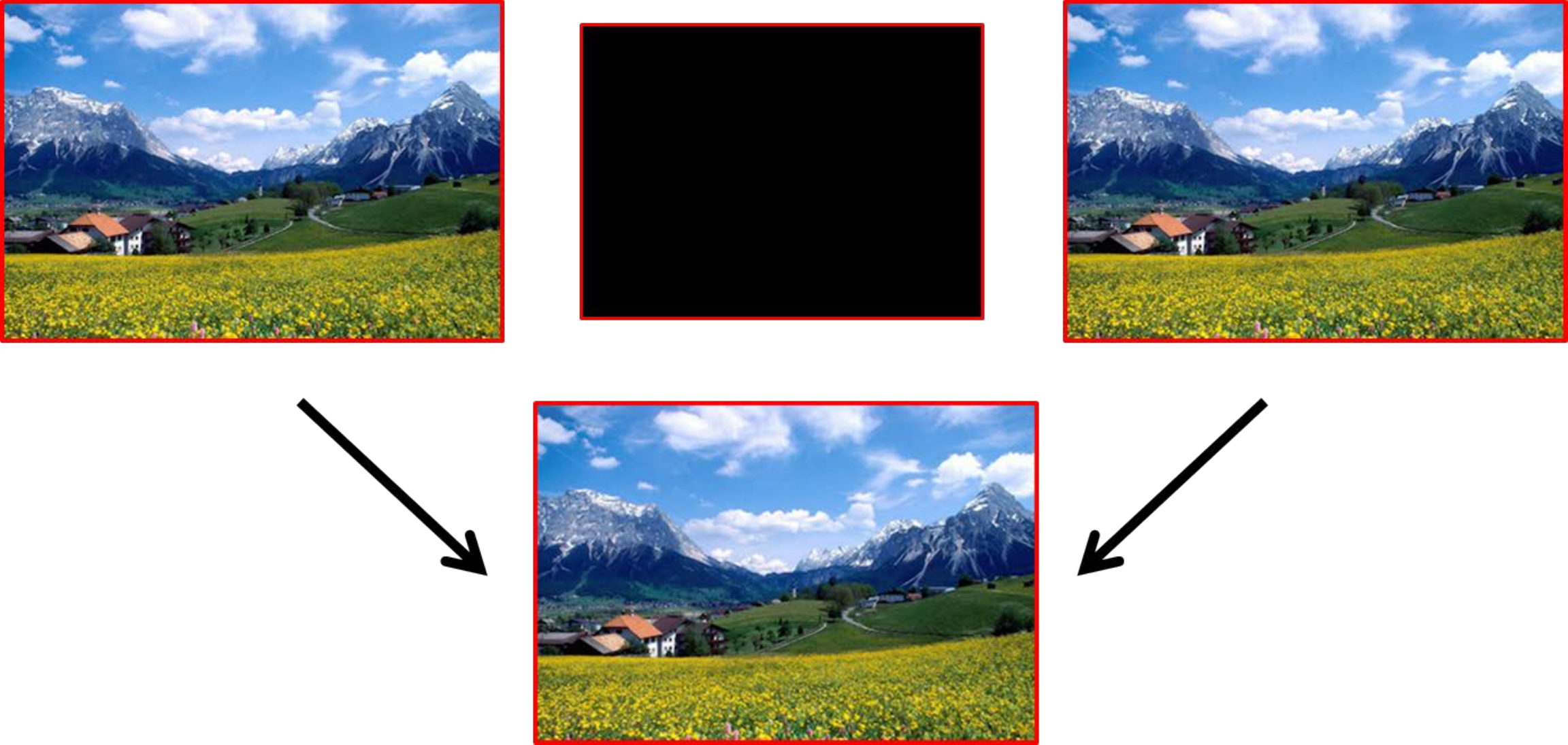}
\caption{A schematic view of how, if we close our eyes for a moment and open them again, we normally merge the before and after views to make one. The landscape here and in Figure \ref{recognition_figure} is from Wallpapers Buzz (\href{http://www.wallpapersbuzz.com/}{www.wallpapersbuzz.com}), reproduced with permission.}
\label{swiss_landscape_figure}
\end{figure}

Here is another example. If, when we are looking at something, we close our eyes for a moment and open them again, what do we see? Normally, it is the same as what we saw before. But recognising that the before and after views are the same, means unifying the two patterns to make one and thus compressing the information, as shown schematically in Figure \ref{swiss_landscape_figure}.

It seems so simple and obvious that if we are looking at a landscape like the one in the figure, there is just one landscape even though we may look at it two, three, or more times. But if we did not unify successive views we would be like an old-style cine camera that simply records a sequence of frames, without any kind of analysis of understanding that, very often, successive frames are identical or nearly so.

\subsection{Recognition}

\begin{figure}[!htbp]
\centering
\includegraphics[width=0.9\textwidth]{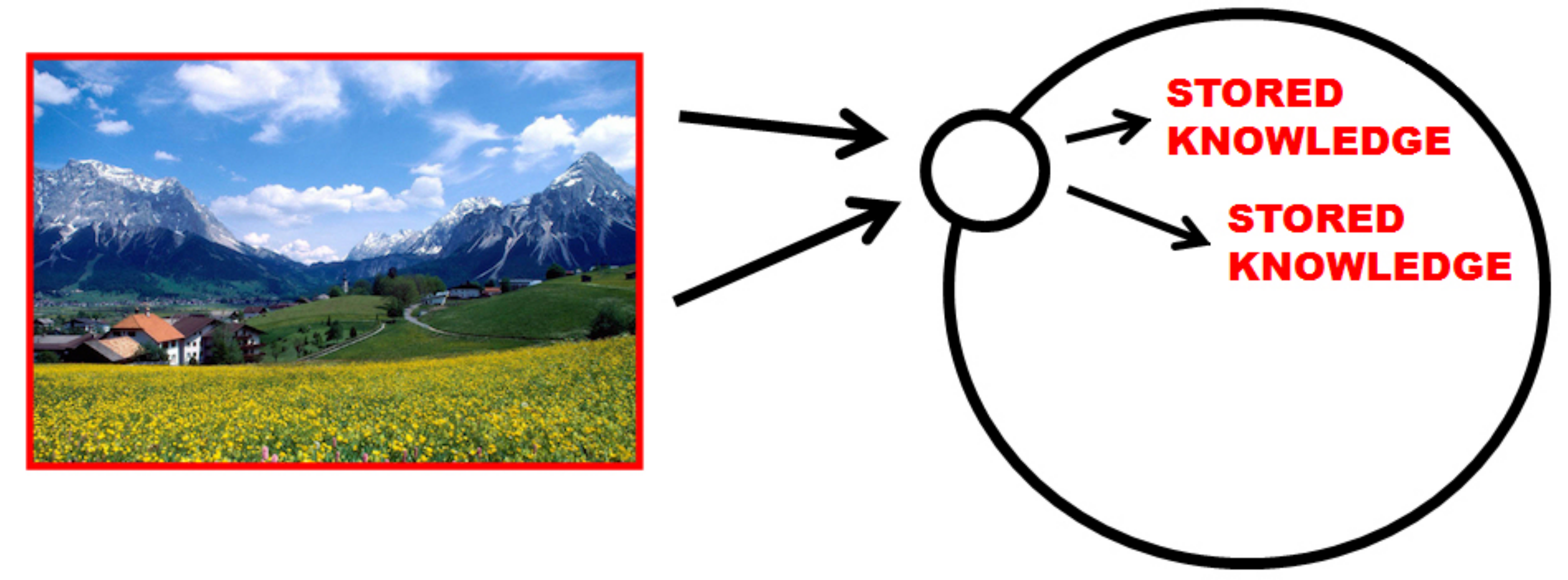}
\caption{Schematic representation of how, in recognition, incoming visual information may be matched and unified with stored knowledge.}
\label{recognition_figure}
\end{figure}

Of course, we can recognise something that we have seen before even if the interval between one view and the next is hours, months, or years. In cases like that, it is more obvious that we are relying on memory, as shown schematically in Figure \ref{recognition_figure}. Notwithstanding the undoubted complexities and subtleties in how we recognise things, the process may be seen in broad terms as one of matching incoming information with stored knowledge, merging or unifying patterns that are the same, and thus compressing the information. If we did not compress information in that way, our brains would quickly become cluttered with millions of copies of things that we see around us---people, furniture, cups, trees, and so on---and likewise for sounds and other sensory inputs.

\section{Binocular vision}

IC may also be seen at work in binocular vision:

\begin{quote}

``In an animal in which the visual fields of the two eyes overlap extensively, as in the cat, monkey, and man, one obvious type of redundancy in the messages reaching the brain is the very nearly exact reduplication of one eye's message by the other eye.'' \cite[p. 213]{barlow_1969}.

\end{quote}

In viewing a scene with two eyes, we normally see one view and not two. This suggests that there is a matching and unification of patterns, with a corresponding compression of information. Evidence in support of that conclusion comes from a demonstration with `random-dot stereograms', as described in \cite[Section 5.1]{sp_vision}.

\begin{figure}[!htbp]
\centering
\includegraphics[width=0.9\textwidth]{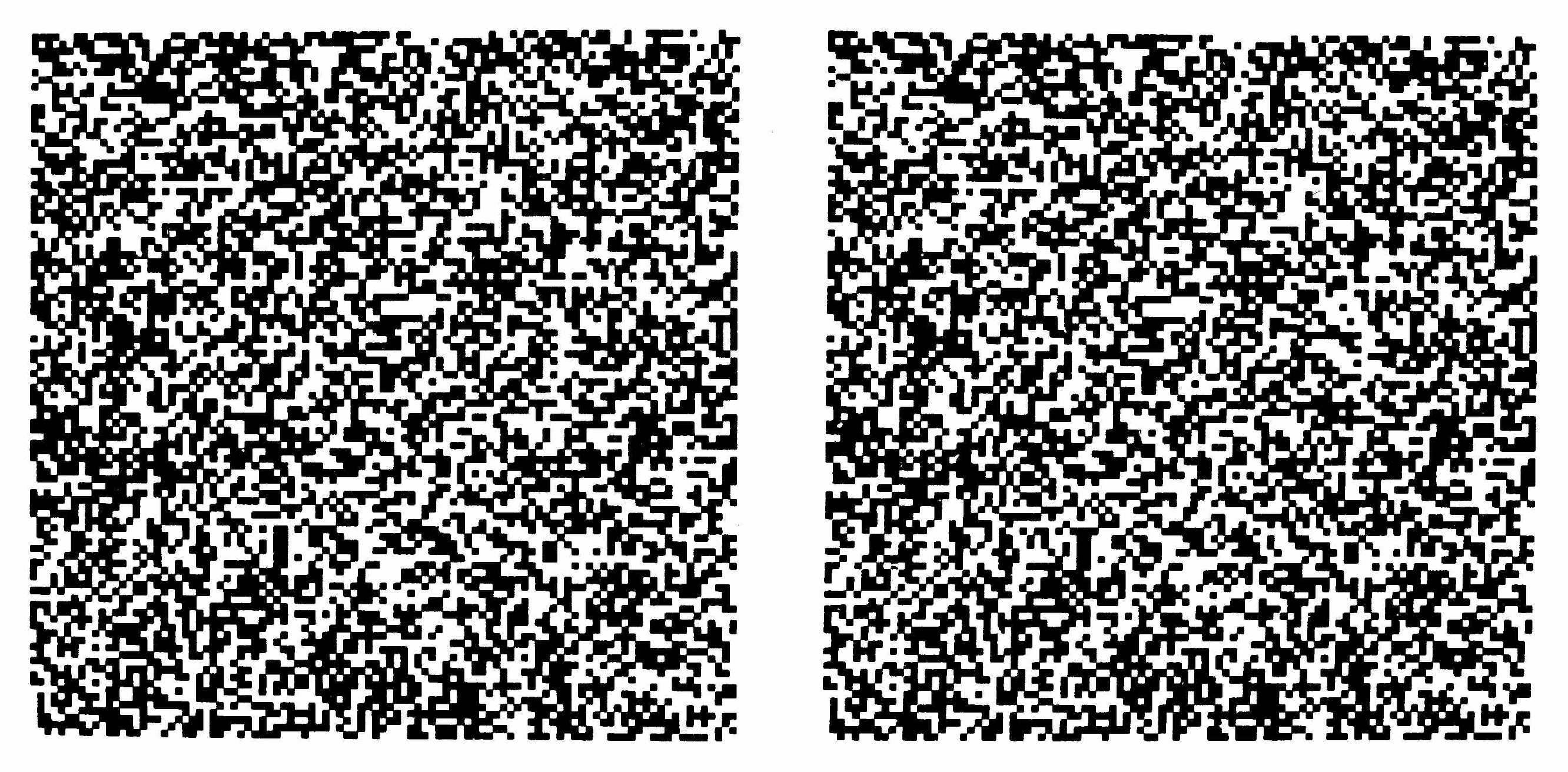}
\caption{A random-dot stereogram from \protect\cite[Figure 2.4-1]{julesz_1971}, reproduced with permission of Alcatel-Lucent/Bell Labs.}
\label{stereogram_1_figure}
\end{figure}

\begin{figure}[!htbp]
\centering
\includegraphics[width=0.9\textwidth]{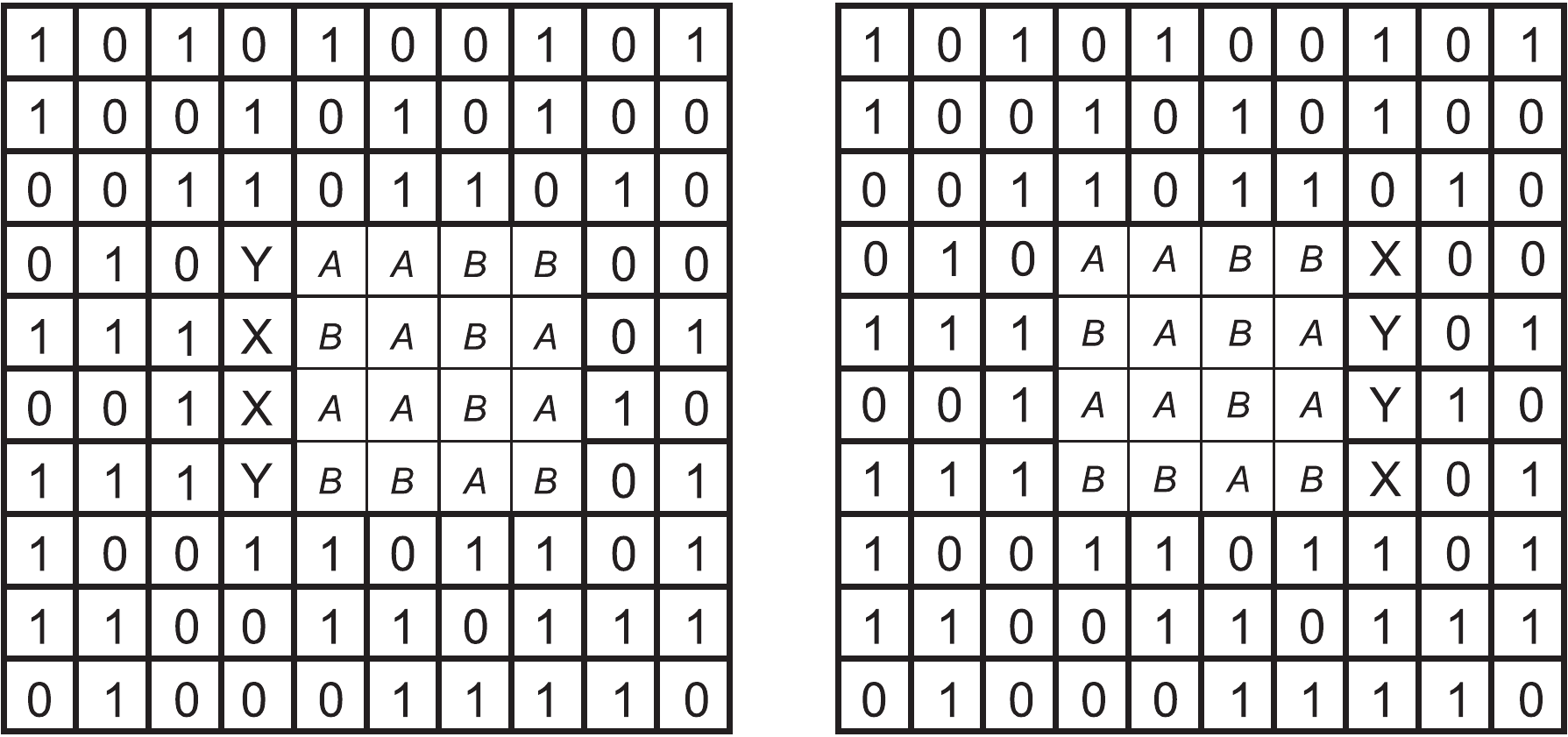}
\caption{Diagram to show the relationship between the left and right images in Figure \ref{stereogram_1_figure}. Reproduced from \protect\cite[Figure 2.4-3]{julesz_1971}, with permission of Alcatel-Lucent/Bell Labs.}
\label{stereogram_2_figure}
\end{figure}

In brief, each of the two images shown in Figure \ref{stereogram_1_figure} is a random array of black and white pixels, with no discernable structure, but they are related to each other as shown in Figure \ref{stereogram_2_figure}: both images are the same except that a square area near the middle of the left image is further to the left in the right image.

When the images in Figure \ref{stereogram_1_figure} are viewed with a stereoscope, projecting the left image to the left eye and the right image to the right eye, the central square appears gradually as a discrete object suspended above the background. Although this illustrates depth perception in stereoscopic vision---a subject of some interest in its own right---the main interest here is on how we see the central square as a discrete object. There is no such object in either of the two images individually. It exists purely in the {\em relationship} between the two images, and seeing it means matching one image with the other and unifying the parts which are the same.

This example shows that, although the matching and unification of patterns is a usefully simple idea, there are interesting subtleties and complexities that arise when two patterns are similar but not identical.

Seeing the central object means finding a `good' match between relevant pixels in the central area of the left and right images, and likewise for the background. Here, a good match is one that yields a relatively high level of IC. Since there is normally an astronomically large number of alternative ways in which combinations of pixels in one image may be aligned with combinations of pixels in the other image, it is not normally feasible to search through all the possibilities exhaustively.

As with many such problems in artificial intelligence, the best is the enemy of the good. Instead of looking for the perfect solution, we can do better by looking for solutions that are good enough for practical purposes. With this kind of problem, acceptably good solutions can often be found in a reasonable time with heuristic search: doing the search in stages and, at each stage, concentrating the search in the most promising areas and cutting out the rest, perhaps with backtracking or something equivalent to improve the robustness of the search. One such method for the analysis of random-dot stereograms has been described by Marr and Poggio \cite{marr_poggio_1979}.

It seems likely that the kinds of processes that enable us to see a hidden object in a random-dot stereogram also apply to how we see discrete objects in the world. The contrast between the relatively stable configuration of features in an object such as a car, compared with the variety of its surroundings as it travels around, seems to be an important part of what leads us to conceptualise the object as an object \cite[Section 5.2]{sp_vision}. Any creature that depends on camouflage for protection---by blending with its background---must normally stay still. As soon as it moves relative to its surroundings, it is likely to stand out as a discrete object.

The idea that IC may provide a means of discovering `natural' structures in the world has been dubbed the `DONSVIC' principle: {\em the discovery of natural structures via information compression} \cite[Section 5.2]{sp_extended_overview}.

\section{Adaptation and run-length coding}

\begin{figure}[!htbp]
\centering
\includegraphics[width=0.9\textwidth]{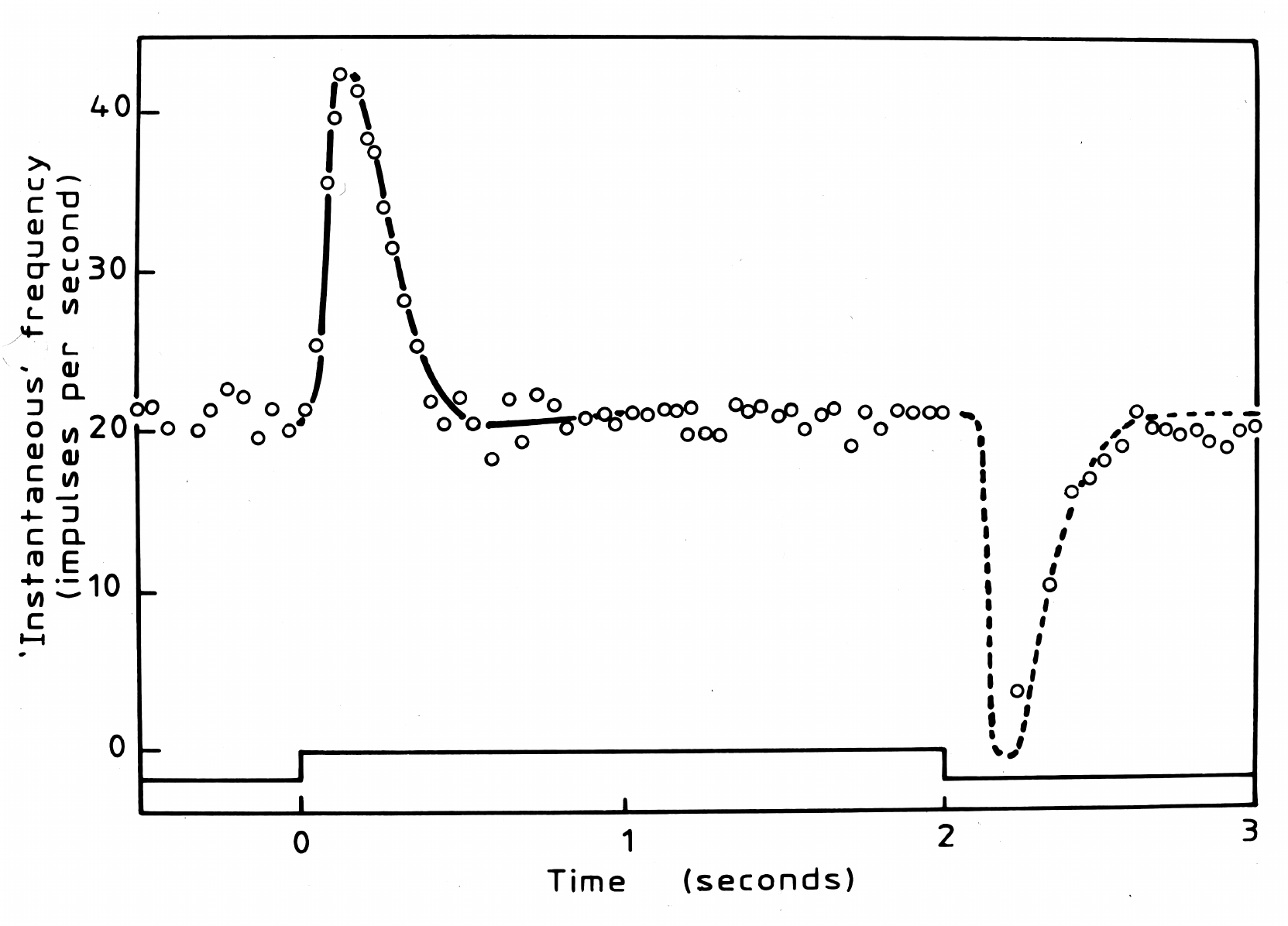}
\caption{Variation in the rate of firing of a single ommatidium of the eye of a horseshoe crab in response to changing levels of illumination. Reproduced from \protect\cite[p.~118.]{ratliff_etal_1963}, with permission from the Optical Society of America.}
\label{limulus_figure}
\end{figure}

IC may also be seen down in the works of vision. Figure \ref{limulus_figure} shows a recording from a single sensory cell ({\em ommatidium}) in the eye of a horseshoe crab ({\em Limulus polyphemus}) as a light is switched on, kept on for a while and then switched off---shown by the step function at the bottom of the figure.

Contrary to what one might expect, the ommatidium fires at a `background' rate of about 20 impulses per second even when it is in the dark (shown at the left of the figure). When the light is switched on, the rate of firing increases sharply but instead of staying high while the light is on (as one might expect), it drops back almost immediately to the background rate. The rate of firing remains at that level until the light is switched off, at which point it drops sharply and then returns to the background level, a mirror image of what happened when the light was switched on.

This pattern of responding---adaptation to constant stimulation---can be explained via the action of inhibitory nerve fibres that bring the rate of firing back to the background rate when there is little or no variation in the sensory input \cite{von_bekesy_1967}. But for the present discussion, the point of interest is that the positive spike when the light is switched on, and the negative spike when the light is switched off, have the effect of marking boundaries, first between dark and light, and later between light and dark.

In effect, this is a form of run-length coding. At the first boundary, the positive spike marks the fact of the light coming on. As long as the light stays on, there is no need for that information to be constantly repeated, so there is no need for the rate of firing to remain at a high level. Likewise, when the light is switched off, the negative spike marks the transition to darkness and, as before, there is no need for constant repetition of information about the new low level of illumination.

It is recognised that this kind of adaptation in eyes is a likely reason for small eye movements when we are looking at something, including sudden small shifts in position (`microsaccades'), drift in the direction of gaze, and tremor \cite{martinez-conde_etal_2013}. Without those movements, there would be an unvarying image on the retina so that, via adaptation, what we are looking at would soon disappear.

Adaptation is also evident at the level of conscious awareness. If, for example, a fan starts working nearby, we may notice the hum at first but then adapt to the sound and cease to be aware of it. But when the fan stops, we are likely to notice the new quietness at first but adapt again and stop noticing it. Another example is the contrast between how we become aware if something or someone touches us but we are mostly unaware of how our clothes touch us in many places all day long. We are sensitive to something new and different and we are relatively insensitive to things that are repeated.

\section{Other kinds of learning}

\begin{figure}[!htbp]
\centering
\includegraphics[width=0.9\textwidth]{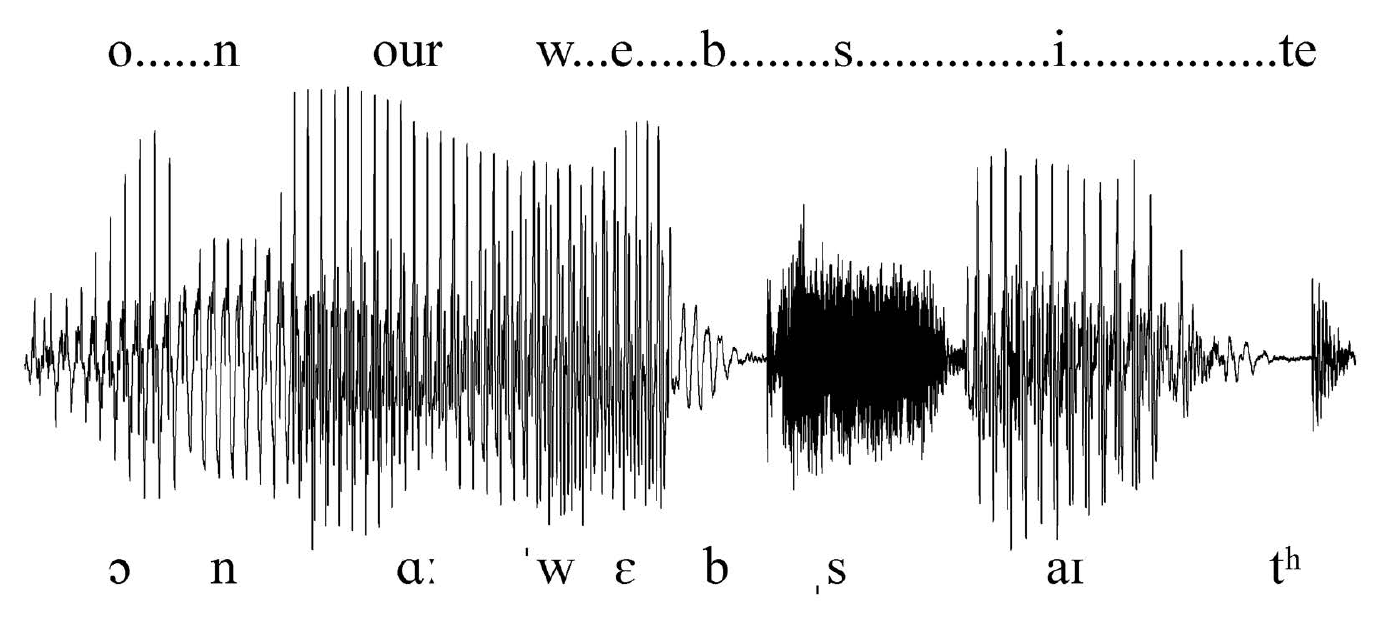}
\caption{Waveform for the spoken phrase ``On our website'' with an alphabetic transcription above the waveform and a phonetic transcription below it. With thanks to Sidney Wood of SWPhonetics (\href{http://swphonetics.com/}{swphonetics.com}) for the figure and for permission to reproduce it.}
\label{speech_waveform_figure}
\end{figure}

As can be seen in Figure \ref{speech_waveform_figure}, people normally speak in `ribbons' of sound, without gaps between words or other consistent markers of the boundaries between words. In the figure---the waveform for a recording of the spoken phrase ``on our website''---it is not obvious where the word ``on'' ends and the word ``our'' begins, and likewise for the words ``our'' and ``website''. Just to confuse matters, there are three places within the word ``website'' that look as if they might be word boundaries.

Given that words are not clearly marked in the speech that young children hear, how do they get to know that language is composed of words? As before, it seems that IC and, more specifically, the DONSVIC principle, provide an answer. It has been shown that, via the matching and unification of patterns, the beginnings and ends of words can be discovered in an English-language text from which all spaces and punctuation has been removed, and this without the aid of any kind of dictionary or other information about the structure of English \cite[Section 5.2]{sp_extended_overview}. It true that there are added complications with speech but it seems likely that similar principles apply.

The DONSVIC principle may also be applied to the process of learning the grammar of a language \cite{wolff_1988}. In addition to the learning of words, the process of grammar discovery or induction includes processes for learning grammatical classes of words (such as nouns, verbs and adjectives) and also syntactic forms such as phrases, clauses and sentences. Ultimately, grammar discovery should also include the learning of meanings and the association of meanings with syntax.

In connection with language learning, IC provides an elegant solution to two problems: {\em generalisation}---how we generalise our knowledge of language without over-generalising; and {\em dirty data}---how we can learn a language despite errors in the examples we hear; with evidence that both these things can be achieved without the correction of errors by parents or teachers. In brief, a grammar that is good in terms of information compression is one that generalises without over-generalising; and such a grammar is also one that weeds out errors in the data. These things are described more fully in \cite[Section 5.3]{sp_extended_overview}.

\section{Perceptual constancies}

It has long been recognised that our perceptions are governed by {\em constancies}:

\begin{itemize}

\item {\em Size constancy}. To a large extent, we judge the size of an object to be constant despite wide variations in the size of its image on the retina \cite[pp.~40-41]{frisby_stone_2010}.

\item {\em Lightness constancy}. We judge the lightness of an object to be constant despite wide variations in the intensity of its illumination \cite[p.~376]{frisby_stone_2010}.

\item {\em Colour constancy}. We judge the colour of an object to be constant despite wide variations in the colour of its illumination \cite[p.~402]{frisby_stone_2010}.

\end{itemize}

These kinds of constancy, and others such as shape constancy and location constancy, may each be seen as a means of encoding information economically. It is simpler to remember that a particular person is ``about my height'' than many different judgements of size, depending on how far away that person is. In a similar way, it is simpler to remember that a particular object is ``black'' or ``red'' than all the complexity of how its lightness or its colour changes in different lighting conditions.

\section{Computing and mathematics}\label{computing_mathematics_section}

If, as seems to be the case, IC is fundamental in our thinking, then it should not be surprising to find that IC is also fundamental in things that we use to aid our thinking: computing in the modern sense where the work is done by machines, and mathematics, done by people or machines. Similar things can be said about logic but the main focus here will be on computing and mathematics, starting with the latter.

\subsection{Mathematics and information compression}\label{mathematics_and_ic_section}

Roger Penrose \cite{penrose_1989} writes:

\begin{quote}

``It is remarkable that {\em all} the SUPERB theories of Nature have proved to be extraordinarily fertile as sources of mathematical ideas. There is a deep and beautiful mystery in this fact: that these superbly accurate theories are also extraordinarily fruitful simply as {\em mathematics}.'' (pp. 225--226, emphasis as in the original).

\end{quote}

\noindent In a similar vein, John Barrow \cite{barrow_1992} writes:

\begin{quote}

``For some mysterious reason mathematics has proved itself a reliable guide to the world in which we live and of which we are a part. Mathematics works: as a result we have been tempted to equate understanding of the world with its mathematical encapsulization.~...~Why is the world found to be so unerringly mathematical?'' (Preface, p.~vii).

\end{quote}

These writings about the ``mysterious'' nature of mathematics, others such as Wigner's \cite{wigner_1960} ``The unreasonable effectiveness of mathematics in the natural sciences'', and schools of thought in the philosophy of mathematics---foundationism, logicism, intuitionism, formalism, Platonism, neo-Fregeanism, and more---have apparently overlooked an obvious point: {\em mathematics can be a very effective means of compressing information}. This apparent oversight is surprising since mathematics is indeed a useful tool in science and, as already mentioned, it is recognised that ``Science is, at root, just the search for compression in the world.'' \cite[p.~247]{barrow_1992}.

Here is an example of how ordinary mathematics---not some specialist algorithm for IC---can yield high levels of IC. Newton's equation for his second law of motion, $s = (gt^2) / 2$,
is a very compact means of representing any realistically-large table of the distance travelled by a falling object ($s$) in a given time since it started to fall ($t$),\footnote{Of course, the law does not work for something like a feather falling in air. The constant, $g$, is the acceleration due to gravity, about $9.8 m / s^2 $.} as illustrated in Table \ref{distance_time_table}. That small equation would represent the table even if it was a 1000 times bigger, or more. Likewise for other equations such as $E = mc^2$, $a^2 + b^2 = c^2$, $P = k / V$, and so on.

\begin{table}[!htbp]
\centering
\texttt{
\begin{tabular}{|r|r|} \hline
\em Distance (m)    & \em Time (sec) \\ \hline
\hline
0.0                 &      0 \\ \hline
4.9                 &      1 \\ \hline
19.6                &      2 \\ \hline
44.1                &      3 \\ \hline
78.5                &      4 \\ \hline
122.6               &      5 \\ \hline
176.5               &      6 \\ \hline
240.3               &      7 \\ \hline
313.8               &      8 \\ \hline
397.2               &      9 \\ \hline
490.3               &     10 \\ \hline
593.3               &     11 \\ \hline
706.1               &     12 \\ \hline
828.7               &     13 \\ \hline
961.1               &     14 \\ \hline
1103.2              &     15 \\ \hline
1255.3              &     16 \\ \hline
{\em Etc}           &    {\em Etc} \\ \hline
\end{tabular}
}
\caption{The distance travelled by a falling object (metres) in a given time since it started to fall (seconds).}
\label{distance_time_table}
\end{table}

In the subsections that follow, we shall dig a little deeper, looking at both mathematics and computing in terms of the ideas outlined earlier (Section \ref{preliminaries_section}): IC via the matching and unification of patterns, chunking-with-codes, schema-plus-correction, and run-lengh coding.

\subsection{Information compression via the matching and unification of patterns}

In mathematics, the matching and unification of patterns can be seen mainly in the matching and unification of names. If, for example, we want to calculate the value of $z$ from these equations: $x = 4$; $y = 5$; $z = x + y$, we need to match $x$ in the third equation with $x$ in the first equation, and to unify the two so that the correct value is used for the calculation of $z$. Likewise for $y$.

The sixth of Peano's axioms for natural numbers---for every natural number $n$, $S(n)$ is a natural number---provides the basis for a succession of numbers: $S(0)$, $S(S(0))$, $S(S(S(0)))$~..., itself equivalent to unary numbers in which $1 = 1$, $2 = 11$, $3 = 111$,~and so on. A numbering system like that is good enough for counting a few things but it is quite unmanageably cumbersome for large numbers. To be practical with numbers of all sizes, the obvious redundancies---in the repetitions of $S$ and of $1$---need to be reduced or eliminated. This can be done via the use of higher bases for numbers---binary, octal, decimal and the like \cite[Section 10.3.2.2]{wolff_2006}.

Emil Post's \cite{post_1943} ``Canonical System'', which is recognised as a definition of `computing' that is equivalent to a Universal Turing Machine, may be seen to work largely via the matching and unification of patterns. Much the same is true of the `transition function' in a Universal Turing Machine.

The matching and unification of patterns may be seen in the way computers retrieve information from computer memory. This means finding a match between the address in the CPU and the address in memory, with implicit unification of the two. It is true that logic gates provide the mechanism for finding an address in computer memory but the process may also be seen as one of searching for a match between the address held in the CPU and the corresponding address in computer memory.

A system like Prolog---a computer-based version of logic---may be seen to function largely via the matching and unification of patterns. Much the same can be said about query-by-example, a popular technique for retrieving information from databases. Other examples will be seen in the subsections that follow.

\subsection{Chunking-with-codes}

If a set of statements is repeated in two or more parts of a computer program then it is natural to declare them once as a `function', `procedure' or `sub-routine' within the program and to replace each sequence with a `call' to the function from each part of the program where the sequence occurred. This may be seen as an example of the chunking-with-codes technique for IC: the function may be regarded as a chunk, with the name of the function as its code or identifier.

In many cases but not all, a name or identifier in computing or in mathematics may be seen to achieve compression of information by serving as a relatively short code for a relatively large chunk of information.

Sometimes, the identifier can be larger than what it identifies but, normally, this can be seen to make sense in terms of IC via schema-plus-correction, next.

\subsection{Schema-plus-correction}

The schema-plus-correction idea may be seen in two main areas: functions with parameters, and object-oriented programming.

\subsubsection{Functions with parameters}

Normally, a function in a computer program, or a mathematical function, has one or more parameters, eg, \texttt{SQRT(number)} (to calculate a square root), \texttt{BIN2DEC(number)} (to convert a binary number into its decimal equivalent, and \texttt{COMBIN(count\_1, count\_2)} (to calculate the number of combinations of \texttt{count\_1} things, taken \texttt{count\_2} at a time).

Any such function may be seen as an example of schema-plus-correction: the function itself may be seen as a chunk of information that may be needed in many different places; the name of the function serves as a relatively short code; and the parameters provide for variations or `corrections' for any given instance.

Imagine how inconvenient it would be if we were not able to specify functions in this way. Every time we wanted to calculate a square root, we would have to write out the entire procedure, and likewise for \texttt{BIN2DEC()}, \texttt{COMBIN()} and the many other functions that people use.

Here we can see why IC may be served, even if an identifier is bigger than what it identifies. Something that is small in terms of numbers of characters, such as the number $9$, may be assigned to the relatively large identifier, ``\texttt{number}'', in \texttt{SQRT(number)}, but that imbalance does little to outweigh the relatively large savings that arise from being able to call the function on many different occasions without having to write it out on each occasion. In any case, the processes of compiling or interpreting a computer program will normally convert long, human-friendly identifiers into short ones that can be processed more efficiently by computers.

\subsubsection{Object-oriented programming}

Apart from functions with parameters, the schema-plus-correction idea is prominent in object-oriented programming. From Simula, through Smalltalk to C++ and beyond, object-oriented languages allow programmers to create software `objects', each one modelled on a `class' or hierarchy of classes. Each such class, which normally represents some real-world category like `person', `vehicle', or `item for delivery', may be seen as a schema. Like a function, each class normally has one or more parameters which may be seen as a means of applying `corrections' to the schema. For example, when a `person' object is created from the `person' class, his or her gender and job title may be specified via parameters.

Classes in object-oriented languages are powerful aids to IC. If, for example, we have defined a class for `vehicle', perhaps including information about the care and maintenance of vehicles, procedures to be followed if there is breakdown outside the depot, and variables for things like engine size and registration number, we avoid the need to repeat that information for each individual vehicle. Attributes of high-level classes are `inherited' by lower-level classes, saving the need to repeat the information in each lower-level class.

\subsection{Run-length coding}

Run-length coding appears in various forms in mathematics, normally combined with other things. Here are some examples:

\begin{itemize}

\item Multiplication (eg, $3 \times 4$) is repeated addition.

\item Division of a larger number by a smaller one (eg, $12 / 3$) is repeated subtraction.

\item The power notation (eg, $10^9$) is repeated multiplication.

\item A factorial (eg, $25!$) is repeated multiplication and subtraction.

\item The bounded summation notation (eg, $\sum_{i = 1}^{5}\frac{1}{i}$) and the bounded power notation (eg, $\prod_{n=1}^{10}\frac{n}{n-1}$) are shorthands for repeated addition and repeated multiplication, respectively. In both cases, there is normally a change in the value of a variable on each iteration, so these notations may be seen as a combination of run-length coding and schema-plus-correction.

\item In matrix multiplication, $AB$ is a shorthand for the repeated operation of multiplying each entry in matrix $A$ with the corresponding entry in matrix $B$.

\end{itemize}

Of course, things like multiplication and division are also provided in programming languages. In addition, there is more direct support for run-length coding with iteration statements like {\em repeat~...~until}, {\em while~...}, and {\em for~...}. For example,

\begin{center}
\begin{BVerbatim}
s = 0;
for (i = 1; i <= 100; i++) s += i;
\end{BVerbatim}
\end{center}

\noindent specifies 100 repetitions of adding $i$ to $s$, with the addition of $1$ to $i$ on each iteration, without the need to write out each of the 100 repetitions explicitly.

Most programming languages also provide for run-length coding in the form of recursive functions like this:

\begin{center}
\begin{BVerbatim}
int factorial(int x)
{
     if (x == 1) return(1) ;
     return(x * factorial(x - 1)) ;
}.
\end{BVerbatim}
\end{center}

\noindent Here, the repeated multiplication and subtraction of the factorial function is achieved economically by calling the function from within itself.

\section{Resolving apparent conflicts}\label{resolving_apparent_contractions_section}

As noted in the Introduction, the idea that IC is fundamental in artificial intelligence, human perception and cognition, and in mainstream computing and mathematics seems to be contradicted by the productivity of the human brain and the ways in which computers and mathematics may be used to create information as well as to compress it; and it seems to be contradicted by the fact that redundancy in information is often useful in both the storage and processing of information. These apparent contradictions and how they may be resolved are discussed briefly here.

\subsection{Decompression by compression}\label{decompression_by_compression_section}

An example of how computers may be used to create information is how the ``hello, world'' message of C-language fame may be printed 1000 times, with a correspondingly high level of redundancy, by a call to `\texttt{hello\_world(1000)}', defined as:

\begin{center}
\begin{BVerbatim}
void hello_world(int x)
{
     printf("hello, world\n");
     if (x > 1) hello_world(x - 1) ;
}.
\end{BVerbatim}
\end{center}

Here, the instruction `\texttt{printf("hello, world$\backslash$n");}' prints a copy of ``hello, world''. Then, when the variable `\texttt{x}' has the value 1000, the next line ensures that the whole process is repeated another 999 times.

The way in which IC may achieve this kind of productivity may be seen via the workings of the SP computer model. When that model \cite[Sections 3.9, 3.10, and 9.2]{wolff_2006} is used to parse a sentence into its constituent parts and sub-parts, as shown in parts (a) and (b) of Figure \ref{fruit_flies_figure}, the model creates a relatively small code as a compressed representation of the sentence \cite[Section 3.5]{wolff_2006}. But exactly the same computer model, using exactly the same processes of IC via the matching and unification of patterns, may reverse the process, reconstructing the original sentence from the code \cite[Section 3.8]{wolff_2006}. This is similar to the way that a suitably-constructed Prolog program may not only be run `forwards' to create `results' from `data' but may also be run `backwards' to create `data' from `results'. A very rough analogy is the way that a car can be driven backwards as well as forwards but the engine is working in exactly the same way in both cases.

Reduced to its essentials, the way that the SP model can be run `backwards' works like this. Using our earlier example, a relatively large pattern like ``\texttt{Treaty on the Functioning of the European Union}'' is first assigned a relatively short code like ``\texttt{TFEU}'' to create the pattern ``\texttt{TFEU Treaty on the Functioning of the European Union}'' which combines the short code with the thing it represents. Then a copy of the short code, ``\texttt{TFEU}'', may be used to retrieve the original pattern via matching and unification with ``\texttt{TFEU}'' within the combined pattern. The remainder of the combined pattern, ``\texttt{Treaty on the Functioning of the European Union}'', may be regarded as the `output' of the retrieval process. As such, it is a decompressed version of the short code. And that decompression has been achieved via a process of IC by the matching and unification of two copies of the short code.

Superficially, using one mechanism to run the model `forwards' and `backwards' has the flavour of a perpetual motion machine: something that looks promising but conflicts with fundamental principles. The critical issue is the size of the short code. It needs to be at least slightly bigger than the theoretical minimum for the process to work as described \cite[Section 3.8.1]{wolff_2006}. If there is some residual redundancy in the code, the SP model has something to work on. With that proviso, ``decompression by compression'' is not as illogical as it may sound.

\subsection{Redundancy is often useful in the storage and processing of information}

There is no doubt that informational redundancy---repetition of information---is often useful. For example:

\begin{itemize}

\item With any kind of database, it is normal practice to maintain one or more backup copies as a safeguard against catastrophic loss of the data.

\item With information on the internet, it is common practice to maintain two or more `mirror' copies in different places to minimise transmission times and to reduce the chance of overload at any one site.

\item The redundancy in natural language can be a very useful aid to comprehension of speech in noisy conditions.

\end{itemize}

These kinds of uses of redundancy may seem to conflict with the idea that IC---which means reducing redundancy---is fundamental in computing and cognition \cite[p.~19]{dodig-crnkovic_2013}. However, the two things may be independent, or the usefulness of redundancy may actually be understood in terms of the SP theory itself.

An example of how the two things may be independent is the above-mentioned use of backup copies of databases: ``... it is entirely possible for a database to be designed to minimise internal redundancies and, at the same time, for redundancies to be used in backup copies or mirror copies of the database~...~Paradoxical as it may sound, knowledge can be compressed and redundant at the same time.'' \cite[Section 2.3.7]{wolff_2006}.

An example of how the usefulness of redundancy may be understood in terms of the SP theory is how, in the retrieval of information from a database or other body of knowledge, there needs to be some redundancy between the search pattern and each matching pattern in the knowledge base (Section \ref{decompression_by_compression_section}). Again, redundancy provides the key to how, in applications such as parsing natural language or pattern recognition, the SP system may achieve good results despite errors of omission, commission or substitution and thus, in effect, suggest interpolations for errors of omission and corrections for errors of commission or substitution \cite[Sections 8 and 9]{sp_extended_overview}, \cite{wolff_sp_intelligent_database}, \cite[Section 6.2]{wolff_2006}.

\section{Conclusion}

This paper presents evidence for the idea that much of artificial intelligence and of human perception and thinking, and much of computing and mathematics, may be understood as compression of information via the matching and unification of patterns.

This is the foundation for the SP theory of intelligence, outlined in Section \ref{outline_sp_theory_section}, with pointers to where further information may be found. The explanatory range of the theory---in perception, reasoning, planning, problem solving, and more---provides indirect support for the idea that IC is an important principle in computing and cognition.

Information compression can mean advantages for creatures: in efficient storage and transmission of information; in being able to make predictions about sources of food, where there may be dangers, and so on; and in corresponding savings in energy. Likewise for artificial systems.

Some aspects of IC and its benefits are so much embedded in our everyday thinking that they are easily overlooked. Most nouns, verbs and adjectives may be seen as short codes for relatively complex concepts, and we frequently create shorthands for relatively long expressions. If we blink or otherwise close our eyes for a moment, we normally merge the before and after views into a single percept. In recognising something after a longer period, we are, in effect, merging the new perception with something that we remember. If we are viewing something with two eyes, we normally merge the two retinal images into a single percept.

IC may be seen in the phenomenon of adaptation in visual perception, in how we learn the structure of words and grammar in language, and in perceptual constancies.

IC via the matching and unification of patterns may be seen in both computing and mathematics. An equation can be a powerful aid to IC. In the processing of computer programs or mathematical equations, IC may be seen in the matching and unification of names. It may also be seen: in the reduction or removal of redundancy from unary numbers to create numbers with bases of 2 or more; in the workings of Post's Canonical System and the transition function in the Universal Turing Machine; in the way computers retrieve information from memory; in systems like Prolog; and in the query-by-example technique for information retrieval.

The chunking-with-codes technique for IC may be seen in the use of named functions to avoid repetition of computer code. The schema-plus-correction technique may be seen in functions with parameters and in the use of classes in object-oriented programming. And the run-length coding technique may be seen in multiplication, in division, and in several other devices in mathematics and computing.

The SP theory resolves the apparent paradox of ``decompression by compression''. And computing and cognition as IC is compatible with the uses of redundancy in such things as backup copies to safeguard data and understanding speech in a noisy environment.

This perspective can be fruitful in research into artificial intelligence, and human perception and cognition including neuroscience \cite{wolff_2006,sp_autonomous_robots,sp_vision,wolff_sp_intelligent_database} and in mainstream computing and its applications \cite{sp_benefits_apps,sp_big_data,wolff_medical_diagnosis}. It may also prove useful in mathematics and its applications.

\bibliographystyle{plain}

\end{document}